\documentclass[runningheads]{llncs}
\usepackage[T1]{fontenc}
\usepackage{graphicx}

\usepackage{epsfig}
\usepackage{amsmath}
\usepackage{amssymb}
\usepackage{subcaption}
\usepackage{booktabs}
\usepackage{braket}
\usepackage{xcolor}
\usepackage{float}
\usepackage{wrapfig}

\usepackage[linesnumbered,ruled,vlined]{algorithm2e}

\SetCommentSty{mycommfont}
\SetKwInput{KwInput}{Input}
\SetKwInput{KwOutput}{Output}

\usepackage[misc]{ifsym}

\usepackage{etoolbox} 
\makeatletter
\patchcmd{\ps@headings}{\rlap{\thepage}}{}{}{}
\patchcmd{\ps@headings}{\llap{\thepage}}{}{}{}
\makeatother
\begin{document}
%

\title{Deconfounded Causality-aware Parameter-Efficient Fine-Tuning for Problem-Solving Improvement of LLMs}

\titlerunning{Deconfounded Causality-aware Parameter-Efficient Fine-Tuning}
%
\author{Ruoyu Wang \inst{1}\textsuperscript{(\Letter)} \and
Xiaoxuan Li \inst{1} \and 
Lina Yao \inst{1,2}}

\authorrunning{Wang et al.}

%

\institute{University of New South Wales \and
Commonwealth Scientific and Industrial Research Organisation, Australia, \email{ruoyu.wang5@unsw.edu.au}, \email{x.l.li@student.unsw.edu.au}, 
\email{lina.yao@unsw.edu.au}}

\maketitle              
\begin{abstract}
Large Language Models (LLMs) have demonstrated remarkable efficiency in tackling various tasks based on human instructions, but studies reveal that they often struggle with tasks requiring reasoning, such as math or physics. This limitation raises questions about whether LLMs truly comprehend embedded knowledge or merely learn to replicate the token distribution without a true understanding of the content. In this paper, we delve into this problem and aim to enhance the reasoning capabilities of LLMs. First, we investigate if the model has genuine reasoning capabilities by visualizing the text generation process at the attention and representation level. Then, we formulate the reasoning process of LLMs into a causal framework, which provides a formal explanation of the problems observed in the visualization. Finally, building upon this causal framework, we propose Deconfounded Causal Adaptation (DCA), a novel parameter-efficient fine-tuning (PEFT) method to enhance the model's reasoning capabilities by encouraging the model to extract the general problem-solving skills and apply these skills to different questions. Experiments show that our method outperforms the baseline consistently across multiple benchmarks, and with only 1.2M tunable parameters, we achieve better or comparable results to other fine-tuning methods. This demonstrates the effectiveness and efficiency of our method in improving the overall accuracy and reliability of LLMs.


\end{abstract}
\section{Introduction}
\label{intro}
Recent years have witnessed remarkable progress on Large Language Models (LLMs) \cite{zhao2023survey}, especially those instruction-following models such as ChatGPT and GPT-4 \cite{openai2023gpt4}. Numerous studies have demonstrated that these models exhibit strong capabilities across a wide range of tasks. However, despite the effectiveness of these models, existing work \cite{jin2023can} shows that they perform poorly on Out-of-Distribution tasks, so fine-tuning with specific tasks and datasets is required to achieve satisfactory results. 

\begin{figure}[t]
  \centering
  \includegraphics[width=0.7\linewidth]{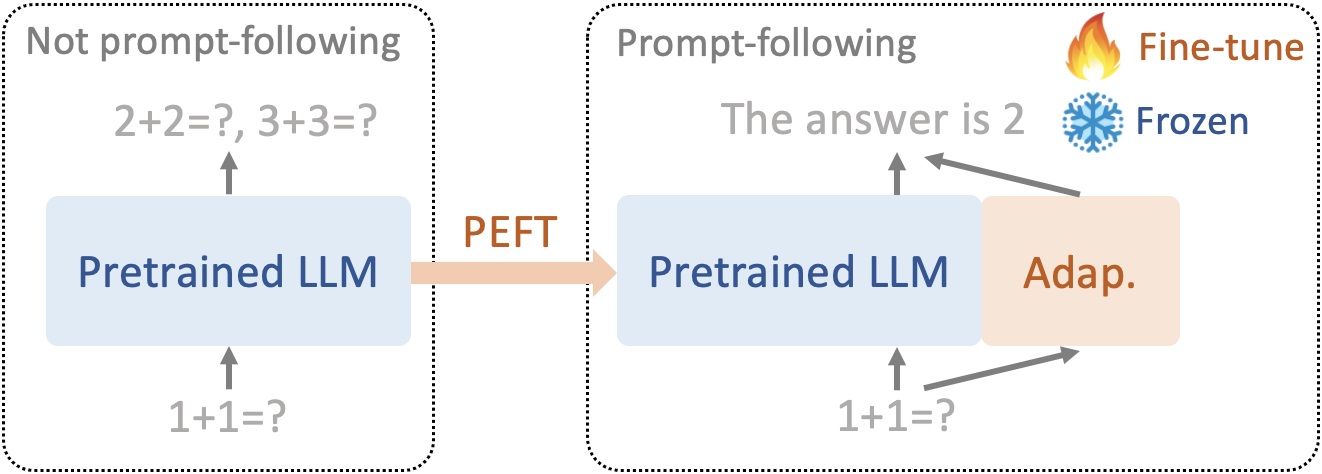}
  \caption{Parameter-Efficient Fine-Tuning (PEFT) methods transform the non-prompt-following model to prompt-following by injecting a small number of learnable parameters into the pre-trained LLM. Our method lies in the domain of PEFT and concentrates on its problem-solving capabilities.}
  \label{fig:peft}
  \vspace{-10pt}
\end{figure}

Nevertheless, fine-tuning large-scale LLMs in full is often prohibitively costly, thus many Parameter-Efficient Fine-Tuning (PEFT) methods have been proposed in recent years, which transform a non-prompt-following model into a prompt-following model by injecting a small number of extra model parameters (Figure~\ref{fig:peft}), thereby greatly decreasing the computational and storage costs. Recent State-of-the-Art PEFT techniques achieve performance comparable to that of full fine-tuning \cite{zhang2023llama,peft}.

While these prompt-following models or fine-tuning methods have been proven to be effective in generating responses based on human instructions, there remains uncertainty regarding whether these models have genuinely acquired knowledge from the text or merely learned the distribution of the word tokens without true comprehension. \cite{wei2022emergent} claimed that the scaling up of language models could significantly enhance their performance, which is usually seen as a piece of evidence that the LLMs can acquire knowledge when it's sufficiently large. However, \cite{schaeffer2023emergent} claims that emergent abilities only appear for specific metrics, and \cite{jin2023can} suggests that these models do not possess any causal reasoning abilities. 

Many discussions have been raised regarding this issue, yet the answer remains inconclusive. Besides, most of these discussions are raised on GPT models, and they are rarely addressed in the context of LLM fine-tuning. Therefore, we investigate this issue in the context of LLM Fine-tuning and propose a novel Parameter Efficient Fine-Tuning (PEFT) method based on Causal Inference techniques to improve the reasoning capabilities of the models. In particular, we first investigate if the model has genuine reasoning capabilities by visualizing the reasoning process at the attention and representation level. Then, we formulate the reasoning process of LLMs into a causal framework, which provides a formal explanation of the problems we observe in the visualization. Finally, we propose Deconfounded Causal Adaptation (DCA), a novel fine-tuning method to improve the model's reasoning capability, and experimentally show the effectiveness and efficiency of our method. The contribution of our paper is three-fold:

\begin{itemize}
    \item We investigate the text generation process of an instruction-following model by visualization in the level of attention and representation, and present empirical evidence that the model lacks genuine causal reasoning capabilities;
    \item We formulate the reasoning process of LLMs in a causal framework, formally explaining the reasons for the observed failure cases in the visualization;
    \item We propose Deconfounded Causal Adaptation (DCA), a novel fine-tuning method to improve the reasoning capability of LLMs, and experimentally demonstrate the effectiveness of our method, which achieves strong performance with only 1.2 Million tunable parameters.
\end{itemize}

\section{Preliminary}
\subsection{LLAMA-Adapter}
\label{pre_adapter}
LLaMA-Adapter \cite{zhang2023llama} is a lightweight adaption method to fine-tune LLaMA into an instruction-following model, which has demonstrated the capability to generate high-quality responses. We conducted our study and built our method based on LLaMA-Adapter due to its effectiveness and efficiency.

\begin{figure}[t]
    \centering
    \begin{minipage}[b]{0.63\textwidth}
        \centering
        \begin{subfigure}[b]{\textwidth}
            \includegraphics[width=\textwidth]{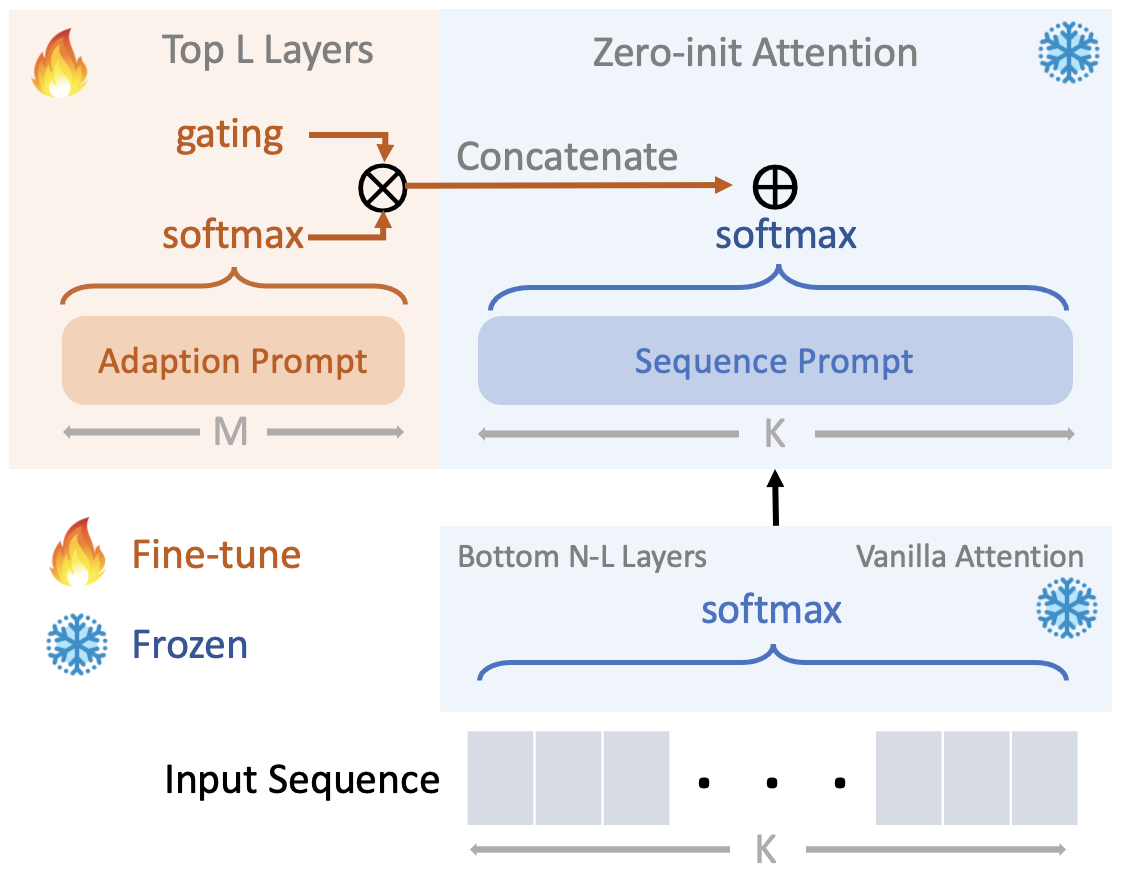}
            \caption{}
            \label{fig:llama_adapter}
        \end{subfigure}
    \end{minipage}
    \begin{minipage}[b]{0.33\textwidth}
        \centering
        \begin{subfigure}[b]{\textwidth}
            \includegraphics[width=\textwidth]{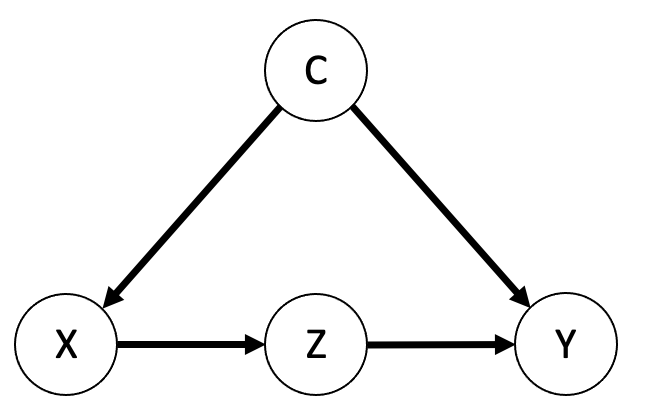}
            \caption{}
            \label{fig:causal_graph1}
        \end{subfigure}
        \vspace{0cm}
        \begin{subfigure}[b]{\textwidth}
            \includegraphics[width=\textwidth]{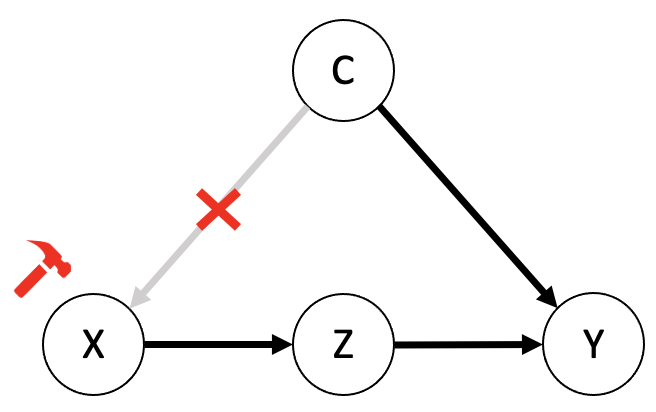}
            \caption{}
            \label{fig:causal_graph2}
        \end{subfigure}
    \end{minipage}
    \caption{(a) The architecture of LLaMA-Adapter. A trainable lightweight adapter is inserted into each of the topmost L layers out of the N transformer layers of LLaMA. Aided by zero-init attention and gating mechanisms, the adaption prompt progressively learns new instructional cues, without disturbing the original pre-trained knowledge; (b) X $\rightarrow$ Z $\rightarrow$ Y is a chain, X $\leftarrow$ C $\rightarrow$ Y is a fork, C $\rightarrow$ Y $\leftarrow$ Z is a collider; (c) We perform intervention $do(X)$ to cut the edge $C \rightarrow X$ so that the causal effect $P(Y|do(X))$ can be estimated.}
    \label{fig:preliminary}
    \vspace{-10pt}
\end{figure}

The architecture of LLaMA-Adapter is illustrated in Figure~\ref{fig:llama_adapter}. For each of the topmost $L$ Transformer layers of LLaMA, an adaption prompt $T_{l} \in \mathbb{R}^{M \times C}$ is concatenated to the original prompt $P_{l} \in \mathbb{R}^{K \times C}$ along the token dimension:
\begin{equation}
    \left[ P_{l}; T_{l} \right] \in \mathbb{R}^{(K+M) \times C}
    \label{eq:adapter_concat}
\end{equation}
where M denotes the length of the adapter to be concatenated, K denotes the original prompt length for each transformer layer, and C denotes the feature dimension of LLaMA’s transformer. This concatenation operation is applied to the corresponding dimension in Key and Value in the self-attention mechanism.

Further, a zero-init attention mechanism with zero gating is proposed to improve the training by injecting the new instructional cues into LLaMA. While calculating the attention score, the softmax function is applied independently to the two components in Equation~\ref{eq:adapter_concat}, and multiplies the concatenated term by a gating factor $g_{l}$, as illustrated in Equation~\ref{eq:adapter_softmax} and Figure~\ref{fig:llama_adapter}.

\begin{equation}
    S_{l}^{g} = \left[ Softmax(S_{l}^{K}); Softmax(S_{l}^{M}) \cdot g_{l} \right]^{T}
    \label{eq:adapter_softmax}
\end{equation}

We highlight components of the LLaMA-Adapter architecture relevant to our method and refer readers to \cite{zhang2023llama} for comprehensive details of this method.

\subsection{Causal Inference}
\label{pre_causal}
In Causality \cite{pearl2009causality}, causal relationships are denoted by Directed Acyclic Graph (DAG). There are three basic building blocks in a causal graph: \textbf{Chain} is the case where one element causally influences another, then leading to the causal impact on a third element, such as X $\rightarrow$ Z $\rightarrow$ Y in Figure~\ref{fig:causal_graph1}. \textbf{Fork} is the case where one element causally influences two other elements, such as X $\leftarrow$ C $\rightarrow$ Y in Figure~\ref{fig:causal_graph1}. \textbf{Collider} is the case where two elements causally influence a third element such as C $\rightarrow$ Y $\leftarrow$ Z in Figure~\ref{fig:causal_graph1}.

\textbf{Confounder} If a variable is the common cause of two other variables, it is called a confounder. Confounders will induce spurious correlations between the two variables, thus disturbing the recognition of the causal effect between them. For example, in Figure~\ref{fig:causal_graph1}, C is a confounder between X and Y. The association between X and Y include the spurious correlations created by the confounder $C$ ($X \leftarrow C \rightarrow Y$) which is non-causal, and the goal of causal inference is to deconfound the spurious correlations so that the true causal relationships between $X$ and $Y$ ($X \rightarrow Z \rightarrow Y$) can be measured.

\textbf{Intervention} In order to measure the causal effect between X and Y, we need to avoid the association flow through the fork  X $\leftarrow$ C $\rightarrow$ Y by blocking the path C $\rightarrow$ X. To this end, we force the variable X = x regardless the value of C. In that case, C no longer affects the value of X and thus path C $\rightarrow$ X is blocked. This process is called intervention in causal inference and is denoted as do(X=x) (Figure~\ref{fig:causal_graph2}). In contrast to $P(Y|X)$, which comprises both causal association and spurious correlations caused by confounder, $P(Y|do(X))$ allows us to measure the genuine causal effect between X and Y.

\section{Our Method}
\subsection{Investigation and Motivation}
\label{method_motivation}
As discussed in Section~\ref{intro}, we aim to investigate if the prompt-following models have genuine causal reasoning capabilities. To this end, we conduct the following experiments. Since models such as ChatGPT and GPT-4 are not available in open-source form, we conduct our study using LLaMA-Adapter \cite{zhang2023llama} to gain access to attention values and representations at each layer.

First, we fine-tune the LLaMA 7B model with LLaMA-Adapter using the \textit{Letter Concatenation} dataset, which will be introduced in Section~\ref{exp_task}. Then, we test the model with two prompts below. The only difference between these two prompts lies in the string within the quotation marks, and as a result, the model answered Prompt A \textbf{correctly}, but \textbf{failed} on Prompt B.

\begin{center}
   \textbf{Prompt A}: \textit{Take the second last letters of the words in “GALLEGOS MORAN” and concatenate them};  
\end{center}

\begin{center}
    \textbf{Prompt B}: \textit{Take the second last letters of the words in “DAVENPORT MAGANA” and concatenate them}. 
\end{center}

To explore the cause of the model's failure on Prompt B, we visualize the attention values in the text generation process by adapting BertViz \cite{vig2019bertviz}, and conduct a thorough comparison between the two test cases on the attention heat map of each attention head across all transformer layers. Consequently, we found that the model's failure on Prompt B can be attributed to the malfunctioning of some particular adapter structures.



\begin{figure}[t]
    \centering
    \begin{minipage}[b]{0.6\textwidth}
        \centering
        \begin{subfigure}[b]{0.5\textwidth}
            \includegraphics[width=\linewidth]{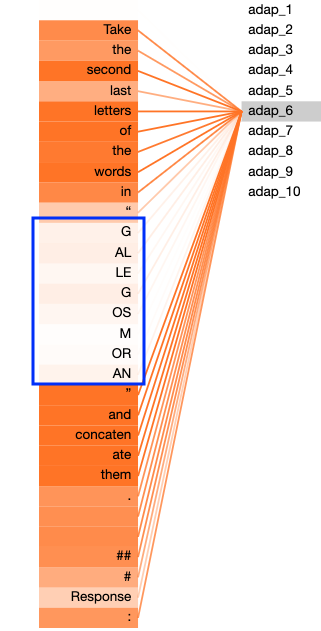}
            \caption{}
            \label{fig:correct_vis}
        \end{subfigure}%
        \begin{subfigure}[b]{0.5\textwidth}
            \includegraphics[width=\linewidth]{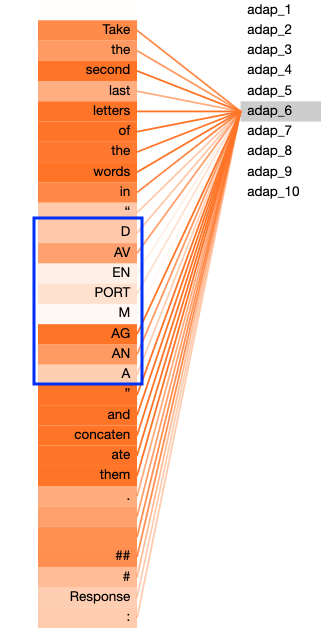}
            \caption{}
            \label{fig:wrong_vis}
        \end{subfigure}
    \end{minipage}
    \begin{minipage}[b]{0.36\textwidth}
        \centering
        \begin{subfigure}[b]{\textwidth}
            \includegraphics[width=\linewidth]{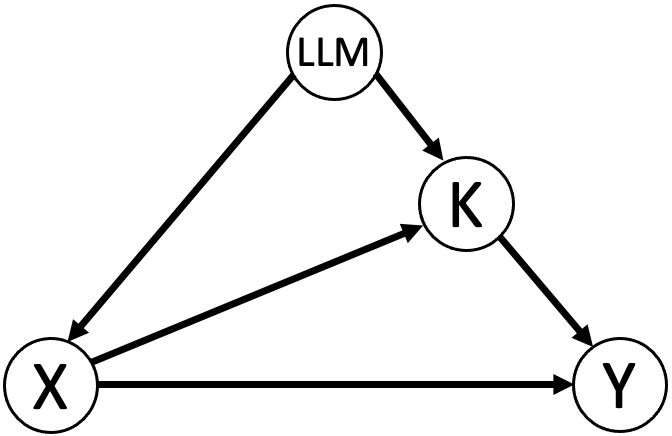}
            \caption{}
            \label{fig:llm_causal_graph1}
        \end{subfigure}
        \par\bigskip
        \begin{subfigure}[b]{\textwidth}
            \includegraphics[width=\linewidth]{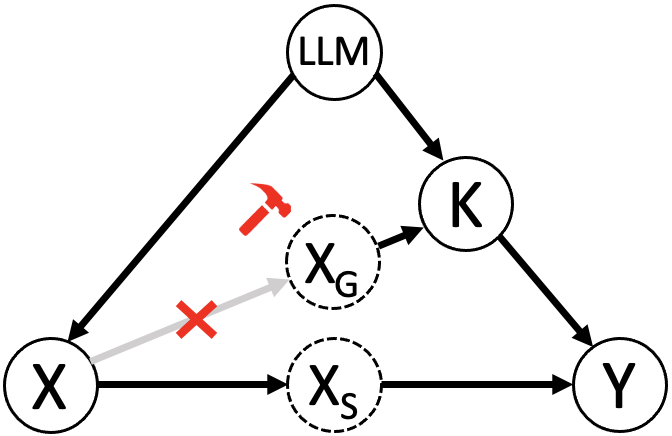}
            \caption{}
            \label{fig:llm_causal_graph3}
        \end{subfigure}
    \end{minipage}
    \caption{(a)-(b) Changing the value of the string affects the functioning of the attention mechanism as highlighted. (c) Causal graph of the reasoning process; (d) We block the backdoor path by performing an intervention on $X_{G}$.}
    \label{fig:vis_n_llm_causal_graph}
\end{figure}

Figure~\ref{fig:correct_vis}-\ref{fig:wrong_vis} provides an example of such malfunctioning structures, where we present the attention values of the sixth element in the adapter ($adap\_6$) located in the 32nd attention head of the last transformer layer of LLaMA-Adapter. We observed that when the model correctly predicts the answer (Figure~\ref{fig:correct_vis}), $adap\_6$ tends to focus on the question rather than the value of the string. However, in Figure~\ref{fig:wrong_vis}, where the model failed to provide the correct answer, it exhibits a focus on a portion of the string, such as token “AG” and “AN” as highlighted. Similar patterns can also be observed in many other cases. Therefore, we empirically conclude that such malfunctioning units are the root cause of the mistake the model made on Prompt B.

In other words, simply replacing the string within the quotation marks significantly affects the \textit{thinking process} of the model. This behaviour starkly contrasts with how humans solve such questions. From a human perspective, Prompt A and Prompt B are nearly identical, if we understand how to solve one of these problems, we inherently possess the method to solve all similar questions. This is because humans understand the world through causal relationships, enabling us to recognize and comprehend the underlying rationales. In contrast, LLMs were constructed based on statistical associations, leading to a deficiency in their capacity to comprehend the question and to do causal reasoning.

Hence, our empirical finding suggests a deficiency in the model's comprehension of the task, as mere string value changes influence the attention mechanism's behaviour. These observations motivate us to enhance the reasoning abilities of these models. Therefore, we introduce our method to improve response quality by fostering the model's capability of causal reasoning. Following this idea, we first formulate the reasoning process of LLMs into a causal framework in Section~\ref{mehtod_spec}, and then propose our causal Fine-tuning method in Section~\ref{method_implementation}.






\subsection{Method Specification}
\label{mehtod_spec}


We formulate the reasoning process of LLMs into a causal framework, as illustrated in Figure~\ref{fig:llm_causal_graph1}. In this framework, X denotes the encoded feature of the prompt, K denotes the relevant knowledge to solve the problem provided by the LLM, and Y denotes the LLM's response to the query.

\textbf{LLM $\rightarrow$ X} When a prompt is presented to the LLM, it encodes the prompt into feature X. Therefore, LLM is the direct cause of X. 

\textbf{LLM $\rightarrow$ K $\leftarrow$ X} Once the prompt is encoded, the LLM offers the relevant knowledge K required to solve the problem in X. Therefore, both the LLM and X are direct causes of K.

\textbf{K $\rightarrow$ Y $\leftarrow$ X} The knowledge K encompasses the method on how to solve the problem described in X, while X contains the question-specific information, such as the values involved in the problem. So both X and K are a cause of Y.

As demonstrated in Section~\ref{method_motivation}, the prompt feature X comprises two independent semantics, one encompasses general problem-solving information, and the other one contains problem-specific information. 
Taking this into consideration, we introduce two additional elements to the graph, namely, the general problem-solving information $X_{G}$, and the problem-specific information $X_{S}$. Both elements are derived from $X$, $X_{G}$ serves as a cause of the problem-solving knowledge K, and $X_{S}$ acts as a mediator between X and Y. 


In this framework, $X_{G}$ and $X_{S}$ should be strictly independent because it's common sense that the problem does not affect the problem-solving skill set. For instance, in the letter concatenation problems, the value of the string within the quotation marks should be independent of the method we use to locate, fetch and concatenate the desired characters. 


However, based on the causal inference theory introduced in Section~\ref{pre_causal}, the independence between $X_{G}$ and $X_{S}$ is not guaranteed. Although there are no direct causal relationships between the two elements, $X$ acts as a confounder between $X_{G}$ and $X_{S}$ and thus creates spurious associations between them. This explains the phenomenon we observed in Figure~\ref{fig:correct_vis}-\ref{fig:wrong_vis}, where altering the value of $X_{S}$ (the string within the quotation marks) affects the reasoning process $X_{G}$ (the functionality of $adap\_6$).

Therefore, to deconfound the spurious association between $X_{G}$ and $X_{S}$, we perform an intervention on $X_{G}$ to block the association from flowing through the path $X_{G} \leftarrow X \rightarrow X_{S}$, as demonstrated in Figure~\ref{fig:llm_causal_graph3}. In that case, changing $X_{S}$ will no longer affect the reasoning process of $X_{G}$.


\subsection{Implementation of Causal Intervention}
\label{method_implementation}
\begin{figure*}[t]
    \includegraphics[width=1\linewidth]{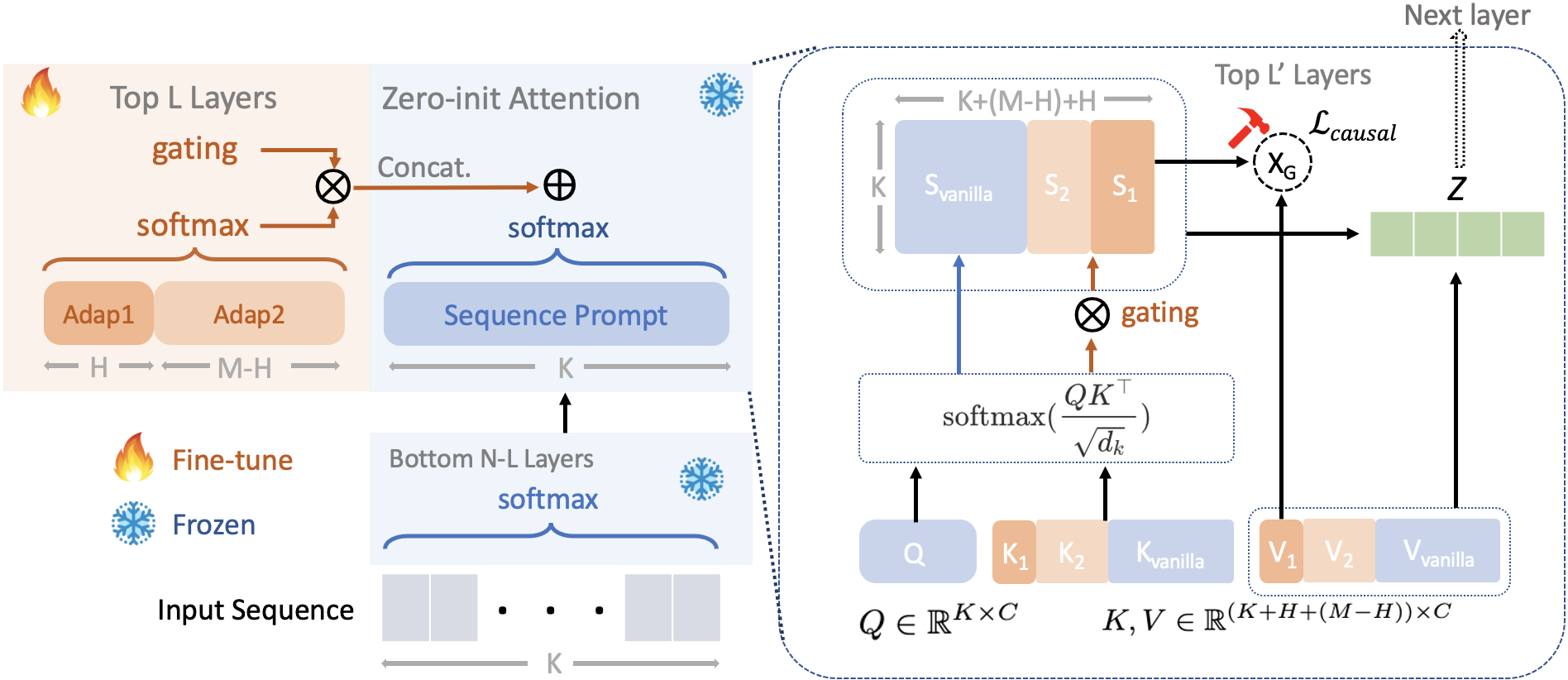}
    \caption{The framework of our method. First, we divide the concatenated Adapter prompt in LLaMA-Adapter into two segments, $Adap1$ and $Adap2$. This affects the dimensions of $K$ and $V$ in the self-attention mechanism, as denoted on the right-hand side. The process of generating the feature $Z$ remains unchanged, and we build our causal loss $\mathcal{L}_{causal}$ by manipulating $Adap1$.}
    \label{fig:framework}
\end{figure*}

In this section, we introduce our method to implement the intervention on $X_{G}$, as illustrated in Figure~\ref{fig:llm_causal_graph3}. First, we assume that the general problem-solving information $X_{G}$ and the problem-specific information $X_{S}$ can be identified by comparison across samples in a dataset, i.e., the differences between data samples are problem-specific, and thus belong to $X_{S}$, and the general problem-solving knowledge, denoted as $X_{G}$, is common across all samples. For instance, in the example given in Section~\ref{method_motivation}, $X_{G}$ contains the method of fetching the desired characters and performing concatenation, and $X_{S}$ contains the order of the characters to be fetched and from which string are these characters to be selected. 

With this assumption, performing the intervention $do(X_{G})$ is equivalent to holding $X_{G}$ invariant across all data samples so that it can maintain the general problem-solving information consistently while changing $X_{S}$. For example, we aim to hold $adap\_6$ invariant across Figure~\ref{fig:correct_vis} and Figure~\ref{fig:wrong_vis}, to avoid it possessing information of $X_{S}$, such as the token “AG” and “AN” in Figure~\ref{fig:wrong_vis}.

Thus, we introduce a causal constraint into the training process to encourage $X_{G}$ to remain invariant across all data samples. Mathematically, we penalize a larger value of variance on $X_{G}$ by introducing a regularization term in Equation~\ref{eq:causal_reg}

\begin{equation}
    \min_{\theta} \mathcal{L}_{CE} + \alpha \mathcal{L}_{causal}
    \label{eq:total_loss}
\end{equation}
\begin{equation}
     \mathcal{L}_{causal} = \mathbb{E}_{l \in L'} \left[ Var(X_{G}) \right]
    \label{eq:causal_reg}
\end{equation}
where $\mathcal{L}_{CE}$ is the Cross-Entropy Loss used to train the token prediction accuracy, and $\alpha$ is the weight of our causal regularization term. We apply this causal regularizer on the topmost $L'$ transformer layers, so we take expectation over these layers, where $L' \leq L$ is a tunable hyper-parameter.

In order to estimate $X_{G}$ in each of the topmost $L'$ layers in Equation~\ref{eq:causal_reg}, we divide the concatenated adapter $T_{l}$ into two separate pieces, $T_{l,1}$ with the length $H$, and $T_{l,2}$ with length $M-H$. Therefore, we rewrite Equation~\ref{eq:adapter_concat} as:



\begin{equation}
    \left[ P_{l};T_{l,1}; T_{l,2} \right] \in \mathbb{R}^{(K+H+(M-H)) \times C}
    \label{eq:adapter_concat_rewrite}
\end{equation}

Similar to the vanilla LLaMA-Adapter, this affects the dimension setting of the Key and Value in the self-attention module. Therefore, we rewrite these two modules as Equation~\ref{eq:K_rewrite} and Equation~\ref{eq:V_rewrite}.

\begin{equation}
    K_{l} = \left[ K_{vanilla}; K_{adap1}; K_{adap2} \right]
    \label{eq:K_rewrite}
\end{equation}

\begin{equation}
    V_{l} = \left[ V_{vanilla}; V_{adap1}; V_{adap2} \right]
    \label{eq:V_rewrite}
\end{equation}

Then, instead of applying the softmax function on the three components independently, we first apply the softmax function on the two original components and multiply with the gating module introduced in the vanilla LLaMA-Adapter, then separate the score matrices into three pieces. Therefore, we have Equation~\ref{eq:S_rewrite}.

\begin{equation}
    S_{l}^{g} = \left[ S_{vanilla}; S_{adap1}; S_{adap2} \right]
    \label{eq:S_rewrite}
\end{equation}

These operations divide the adapter architecture into two segments. Then we treat these two segments as $X_{G}$ and $X_{S}$ respectively, enabling us to impose distinct constraints on each of them. In particular, we treat $T_{l,1}$ with length $H$ as the section controlling the general problem-solving information $X_{G}$. Therefore, $X_{G}$ can be estimated by Equation~\ref{eq:xg_estimate}.
\begin{equation}
    X_{G} \approx S_{adap1} \cdot V_{adap1}
    \label{eq:xg_estimate}
\end{equation}

Finally, we aggregate this quantity in each of the topmost $L'$ layers and take expectation to form the causal regularizer as introduced in Equation~\ref{eq:causal_reg}. The architecture of our method is illustrated in Figure~\ref{fig:framework}. The modules involved in the calculation of $\mathcal{L}_{causal}$ are coloured in dark red.





\section{Experiment}
\label{exp}

\subsection{Experimental Settings}
\label{exp_setting}
We build our method by fine-tuning LLaMA 7B model \cite{touvron2023llama}, thus all the parameters related to dimensions and layers remain unchanged, such as the number of transformer layers is 32, and each transformer layer has 32 attention heads. Also, the feature dimension is 128 for each attention head, thus the total feature dimension is 4096. We train the model with a maximum sequence length of 256 and use AdamW for optimization with a learning rate equal to 1e-3. All the models are fine-tuned for 5 epochs with a batch size of 4 for a fair comparison.

In terms of the parameters introduced by vanilla LLaMA-Adapter, we set $L=20$ and $M=10$, which means we fine-tune the top 20 transformer layers by appending an adapter prompt of length 10 on each of them. For the parameters $H$ and $\alpha$ introduced by our method, we set $H$ as 2 and $\alpha$ as 1 in all experiments. The parameter $L'$ is data-dependent, and we use 20 for Letter Concatenation, 10 for Date Understanding, 3 for AddSub and Math10k, and 1 for Math401. All other settings, if not specified here, remain the same as in \cite{zhang2023llama}.

\subsection{Tasks for Evaluation}
\label{exp_task}
We evaluate the performance of our method by three types of reasoning tasks:

\textbf{Symbolic Reasoning} We construct a more challenging version of the last letter concatenation problem in \cite{wei2022chain} because the models could almost perfectly solve the problems if the models are fine-tuned with it. Therefore, we ask the model to perform \textit{second last} letter concatenation, such as \textit{Take the second last letters of the words in ``Lady Gaga" and concatenate them}.

\textbf{Commonsense Reasoning} We test the models with Date Understanding data \cite{srivastava2022beyond}, where each data sample asks a multiple-choice question such as \textit{If today is Jan 1, 2023, what day is tomorrow in MM/DD/YYYY?}

\textbf{Arithmetic Reasoning} We test the models on three datasets, \textbf{Math401} \cite{yuan2023large}, which comprises basic arithmetic questions such as \textit{1+2=?}, \textbf{AddSub} \cite{hosseini2014learning} and \textbf{Math10k} \cite{hu2023llm}, both comprises math word questions such as \textit{Tom found 7 seashells but 4 were broken . How many unbroken seashells did Tom find?}.

\begin{table*}[t]
  \centering
  \small
  \caption{Accuracies of models based on LLaMA-7B. Our method achieves better or comparable results to other methods with only 1.2M tunable parameters.}
  \begin{tabular}{l | c | l l l l l | l }
    \toprule
    & Params & LConcat & Date & Math401 & AddSub & Math10k & Avg.\\
    \midrule
    
    Alpaca-7B \cite{taori2023stanford} & - & 0.0 & 52.2 & 9.8 & 22.3 & 10.2 & 18.9 \\
    Vicuna-7B \cite{chiang2023vicuna} & - & 0.0 & 29.4 & 29.2 & 38.6 & 15.3 & 28.5 \\
    Koala-7B \cite{geng2023koala} & - & 0.0 & 54.3 & 25.6 & 32.4 & 12.7 & 25.0 \\
    Baize-7B \cite{xu2023baize} & - & 0.0 & 44.9 & 28.3 & 34.4 & 11.2 & 23.8 \\
    LLaMA2-7B ct. \cite{touvron2023llama2} & - & 0.0 & 56.8 & 30.6 & 56.7 & 21.6 & 33.1 \\
    Mistral-7B-ins. \cite{jiang2023mistral} & - & 0.0 & 54.6 & 28.6 & 39.6 & 13.2 & 27.2 \\

    \midrule
    
    S-Adapter$^{h}$ \cite{houlsby2019parameter} & 134M & 80.1 & 79.8 & 20.2 & 78.1 & 29.9 & 57.6 \\
    S-Adapter$^{p}$ \cite{pfeiffer2020mad} & 68M & 77.3 & 79.3 & 21.5 & 82.1 & 24.1 & 56.9 \\
    P-Adapter \cite{he2021towards} & 200M & 80.4 & 82.2 & 22.1 & 84.7 & 29.5 & 59.8 \\
    LoRA \cite{hu2021lora} & 4.2M  & 80.8 & 82.6 & 23.6 & 83.3 & 30.8 & 60.2 \\
    AdaLoRA \cite{zhang2023adaptive} & 3.8M & 80.9 & \underline{83.0} & 23.4 & \underline{85.4} & 30.9 & \underline{60.7} \\
    Prefix-Tune \cite{li2021prefix} & 7.0M & 80.2 & 78.3 & \textbf{25.2} & 57.0 & \underline{34.9} & 55.3 \\
    Prompt-Tune \cite{lester2021power} & 2.0M & 78.3 & 82.6 & 21.2 & 62.3 & 24.7 & 53.8 \\
    KronA \cite{edalati2022krona} & 4.2M & \underline{81.7} & 82.8 & 23.4 & 83.0 & 31.4 & 60.5 \\
    LoftQ \cite{li2023loftq} & 4.0M & 80.9 & 82.7 & 23.0 & 83.7 & 29.8 & 60.0 \\
    
    \midrule
    
    LLaMA-Adap. & 1.2M  & 75.3 & 78.3 & 21.6 & 83.6 & 30.2 & 57.8 \\ 					
    DCA (Ours) & 1.2M & \textbf{82.1}{\fontsize{6}{7}\selectfont \textcolor{teal}{(+6.8)}} & \textbf{84.7}{\fontsize{6}{7}\selectfont \textcolor{teal}{(+6.4)}} & \underline{24.6}{\fontsize{6}{7}\selectfont \textcolor{teal}{(+3.0)}} & \textbf{86.3}{\fontsize{6}{7}\selectfont \textcolor{teal}{(+2.7)}} & \textbf{35.3}{\fontsize{6}{7}\selectfont \textcolor{teal}{(+5.1)}} & \textbf{62.6}\\
    
    \bottomrule
  \end{tabular}
  \label{tab:rst}
\end{table*}

\subsection{Baselines and Comparison Methods}
\label{exp_baseline}
We compare our method with other methods from three perspectives to conduct a comprehensive comparison:

1) We compare our method with the vanilla \textbf{LLaMA-Adapter} \cite{zhang2023llama}. Since we build our method based on LLaMA-Adapter, this comparison allows us to understand the direct impact of implementing our method. All common settings between the two methods such as parameters are kept the same to ensure a fair comparison. The results of this comparison is presented in the \textbf{bottom block} of Table~\ref{tab:rst}, and we highlight the margin achieved by our method in green.

2) We compare our method with the other parameter-efficient fine-tuning (PEFT) methods, as listed in the \textbf{middle block} of Table~\ref{tab:rst}. We apply these methods on LLaMA 7B, and the results are obtained with the library and hyper-parameters provided by \cite{peft,hu2023llm}. We present the results and the number of learnable parameters allowing us to compare our method with the baseline methods in terms of both effectiveness and efficiency. 

3) We compare our method with several pre-trained prompt-following models with the size of 7B, as listed in the \textbf{top block} of Table~\ref{tab:rst}. These models do not lie in the domain of PEFT and thus are not directly comparable to our method. They are either obtained by full fine-tuning or pre-trained with massive conversational data. We compare our method with these models to investigate their performances on the reasoning tasks and evaluate if task-specific fine-tuning is necessary to achieve satisfactory results.

\subsection{Overall Results}
\label{exp_results}
The results are presented in Table~\ref{tab:rst}, where the numbers denote the accuracies the methods achieve on each dataset. While comparing our method with the three types of baselines outlined above, our findings also fall into three aspects:

1) \textbf{Compared with LLaMA-Adapter}: Our method consistently outperforms LLaMA-Adapter by a considerable margin on all datasets, as highlighted in green in Table~\ref{tab:rst}. Since all the common settings of the two methods remain the same, the results directly demonstrate the impact of our causal method.

2) \textbf{Compared with the other PEFT methods}: We found that while the vanilla LLaMA-Adapter does not always outperform the baseline methods, our method, in contrast, achieves either the highest or the second highest score across all datasets. Even though a few methods may perform better than our method on some particular datasets, it is worth noting that our method has only 1.2M learnable parameters, which is the least among all methods. In summary, our method achieves better or comparable results with other PEFT methods, with much less learnable parameters.

3) \textbf{Compared with pre-trained models}: We found that the performance of pre-trained models is generally not satisfactory compared with the PEFT methods. While these models achieve fair performances on some datasets, they face significant challenges in the LConcat task. Notably, it was observed that none of the pre-trained models under consideration could accurately respond to the Letter Concatenation questions. To ensure this phenomenon is not due to the bias in our prompt, we endeavoured to rephrase the questions in LConcat, however, the models consistently exhibited an inability to comprehend the prompts and frequently provided irrelevant or meaningless responses. We speculate that this is due to the insufficient inclusion of training data of this specific nature during the model's fine-tuning phases.

\textbf{Summary} Our experiments suggest that fine-tuning on specific tasks is necessary to achieve satisfactory results. And, among the Parameter-Efficient Fine-Tuning methods, our method achieves \textbf{better or comparable} results with \textbf{much less learnable parameters} and computational resources.

\subsection{Effects of New Parameters}
To further investigate the mechanism of our method, we study the impact of parameters introduced by our method, namely, the length $H$ of adaption prompts to be treated as $X_{G}$, the weight $\alpha$ of the regularization term  $\mathcal{L}_{causal}$, and the number of layers $L'$ to be used to calculate $\mathcal{L}_{causal}$. 

\begin{figure}[t]
    \centering
    \begin{subfigure}[b]{0.45\textwidth}
        \includegraphics[width=\linewidth]{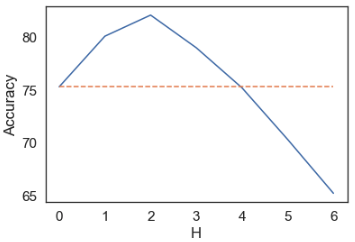}
        \caption{Effect of $H$}
        \label{fig:param_h_effect}
    \end{subfigure}%
    \begin{subfigure}[b]{0.45\textwidth}
        \includegraphics[width=\linewidth]{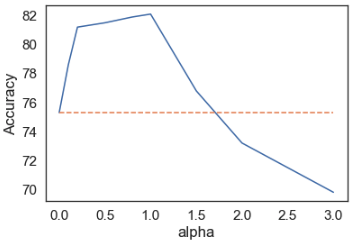}
        \caption{Effect of $\alpha$}
        \label{fig:param_alpha_effect}
    \end{subfigure}
    \caption{Effect of $H$ and $\alpha$ on LConcat. The red dot line denotes baseline accuracy. The value of these parameters should be chosen carefully, otherwise may harm the performance when the values are too large.}
    \label{fig:param_effect_vis}
\end{figure}

\textbf{Choice of $H$ and $\alpha$} We visualize the effect of $H$ and $\alpha$ on the Letter Concatenation dataset in Figure~\ref{fig:param_h_effect} - \ref{fig:param_alpha_effect}, where the x-axis denotes the value of the parameters, and the y-axis denotes the accuracy obtained by the model. Similar trends can be observed in both charts that increasing the value of $H$ and $\alpha$ can improve the performance of the model, but excessive values can be detrimental. This aligns with our intuition. For $H$, if a substantial fraction of the adapter remains fixed as $X_{G}$, then only a limited part of the adapter could be left to address $X_{S}$, which compromises its efficacy in managing problem-specific information. For $\alpha$, if a large weight is employed for $\mathcal{L}_{causal}$, the module to handle $X_{G}$ might remain constant and cannot encode any information.

\textbf{Choice of $L'$} We found the optimal choice of $L'$ is data-dependent. On datasets like Letter Concatenation, where all the prompts follow the same format, a larger $L'$ is beneficial to the performance. In contrast, on datasets like AddSub, where the questions are not necessarily in the same template, a smaller $L'$ is preferable. This is intuitively reasonable, because for those datasets where the prompts are close enough in the first place, encouraging the model to extract $X_{G}$ from the bottom layers grants us more control over the reasoning process. In contrast, for those datasets where the prompts are not sufficiently close, $X_{G}$ can only be extracted and controlled when the representations have been aggregated to a certain level. In that case, a large $L'$ would limit the model's potential for aggregating the high-level information.



\subsection{Further Discussions}
\label{exp_discussion}
\textbf{Applicable scenarios} 
We illustrate the motivation and idea of our method in Section~\ref{method_motivation}. However, it is worth noting that our method is not limited to the case of the same pattern questions. Instead, prompts in different formats also benefit from our method. As demonstrated in Section~\ref{exp}, our method benefits a wide range of reasoning tasks with various datasets. This is because we encourage the model to extract the ``approach'' of solving problems. In other words, as long as a prompt involves reasoning, there will be some problem-solving skills ($X_{G}$), and our method is applicable to the scenario. For example, in date understanding and math word questions, where the prompts vary significantly, our method still benefits the performance as illustrated in Table~\ref{tab:rst}, because we encourage the model to extract the high-level knowledge, such as the meaning of ``tomorrow'', ``end of the month'' or the math operations such as ``Add'', ``Subtract'', and keep these problem-solving skills invariance across all data samples. In contrast, our method does not apply to the general Q\&A questions, such as \textit{Tell me about Alpaca}, because these questions do not require reasoning capabilities and there is no ``approach'' to answer these questions.

\textbf{Few-shot experiments} Few-shot prompt method such as Chain-of-Thought (COT) \cite{wei2022chain} is known to be useful on large models like ChatGPT/GPT4, but it does not apply to PEFT methods, so we did not include these experiments in our paper. To elaborate, COT works well on ChatGPT/GPT4 because those models are fine-tuned by a massive amount of prompt-answer pairs with one-shot examples, enabling the model to utilize one-shot information effectively. In contrast, our method fine-tunes a non-prompt-following LLMs (LLaMA) with task-specific data aiming for improved performance on the task. Since the data does not contain any one-shot prompts, the model will not be able to utilize the one-shot information. As a matter of fact, our experiments reveal that COT is even \textbf{harmful} to the result in such cases.

\textbf{Finetuning a prompt-following model} We also conduct experiments to apply our method on prompt-following models such as Alpaca. As a result, it achieves an accuracy of 75.3 on LConcat, and 79.8 on Date Understanding datasets, which is not comparable to the result we achieved using the original non-prompt following LLaMA. We speculate this is because such instruction-tuned LLMs (such as Alpaca/Vicuna) are also based on the original foundation model such as LLaMA, and it has been fine-tuned with the data that are not closely related to our downstream tasks, thus dropping some information relevant to our task, thus harming the performance. Therefore, we empirically conclude that it would be a better practice to fine-tune the foundation model, rather than an existing instruction-following model.


\section{Related Works}
{\bf Reasoning in LLMs}. Instruction-following LLMs have been employed on many tasks involving reasoning recently, including but not limited to Mathematics, Logical Reasoning, and Symbolic Reasoning \cite{qiao2022reasoning,wei2022chain,zhao2023survey}. Many of these methods investigate LLM's reasoning capabilities from its output using Chain-of-Thought prompting strategy \cite{wei2022chain}. Apart from these, some works build thinking pipelines \cite{besta2023graph,yao2023tree} to achieve the final goal step-by-step.

{\bf Causal Inference in Machine Learning}. 
Causal inference has been applied to many vision tasks in recent years such as image recognition \cite{yue2020interventional,tang2020long} and Image Generation \cite{kocaoglu2017causalgan}. These works first construct causal graphs to explain the task, then use causal inference methods to eliminate the spurious association and improve the performance of the models. Besides, causal inference techniques are also used in Representation Learning \cite{shen2022weakly,yang2021causalvae}.

\subsection{Relationships with our method}
Existing works typically discuss LLMs' reasoning abilities based on their input and output \cite{qiao2022reasoning}. However, we argue that solving causality-related tasks or providing the thinking processes by words do not necessarily indicate the model's reasoning capability, because simply mimicking token distribution could achieve equivalent outcomes. Our work, in contrast, discusses the reasoning capabilities of LLMs in the level of attention and representation, thus offering a novel perspective on this matter. Besides, the novelty of our method also involves applying causality in LLM fine-tuning, which was rarely discussed in earlier literature.

\section{Conclusion}
In this paper, we first investigated the reasoning capabilities of the prompt-following LLMs by visualizing the attention values in the thinking process, and empirically suggest that these models lack genuine causal reasoning capabilities. Then, we formulate the reasoning process of LLMs into a causal inference framework to explain the issues observed in the visualization. Finally, we propose Deconfounded Causal Adaptation (DCA), a causal fine-tuning method to improve the model's reasoning capability. Experiments show our method effectively enhances the reasoning capabilities of the models and outperforms baseline methods consistently. Besides, we also discuss the applicable scenarios of our method and analyze the effect of our method with different settings thoroughly.

\bibliographystyle{splncs04}
\bibliography{reference}

\end{document}